\documentclass{article}
\usepackage{graphicx} 
\usepackage{booktabs}
\usepackage{multirow}
\usepackage{amsmath}
\usepackage{amssymb}
\usepackage{algorithm}
\usepackage{algorithmic}
\usepackage{url}
\usepackage[T1]{fontenc}
\usepackage{hyperref}
\usepackage{authblk}
\newcommand{\cmark}{\ding{51}}  
\newcommand{\xmark}{\ding{55}}  
\usepackage{pifont} 
\title{VP-Hype: A Hybrid Mamba-Transformer Framework with Visual-Textual Prompting for Hyperspectral Image Classification}
\author[1,2]{Abdellah Zakaria Sellam}
\author[3]{Fadi Abdeladhim Zidi}
\author[4]{Salah Eddine Bekhouche}
\author[5]{Ihssen Houhou}
\author[7]{Marouane Tliba}
\author[1,2]{Cosimo Distante}
\author[6]{Abdenour Hadid}

\affil[1]{Institute of Applied Sciences and Intelligent Systems (ISASI), CNR, 73100 Lecce, Italy}
\affil[2]{Dept. of Engineering for Innovation, University of Salento, 73100 Lecce, Italy}
\affil[3]{VSC Laboratory, Department of Electronics and Automation, University of Biskra, Biskra, Algeria}
\affil[4]{Dept. of Computer Science and AI, University of the Basque Country (UPV/EHU), Spain}
\affil[5]{VSC Laboratory, Department of Electronics and Automation, University of Biskra, Algeria}
\affil[6]{Sorbonne Center for Artificial Intelligence, Sorbonne University Abu Dhabi, Abu Dhabi, UAE}
\affil[7]{Institut Galilée, Université Sorbonne Paris Nord, F-93430, Villetaneuse, France}
\begin{document}

\maketitle
\begin{abstract}
Accurate classification of hyperspectral imagery (HSI) is often frustrated by the tension between high-dimensional spectral data and the extreme scarcity of labeled training samples. While hierarchical models like LoLA-SpecViT have demonstrated the power of local windowed attention and parameter-efficient fine-tuning, the quadratic complexity of standard Transformers remains a barrier to scaling. We introduce VP-Hype, a framework that rethinks HSI classification by unifying the linear-time efficiency of State-Space Models (SSMs) with the relational modeling of Transformers in a novel hybrid architecture. Building on a robust 3D-CNN spectral front-end, VP-Hype replaces conventional attention blocks with a Hybrid Mamba-Transformer backbone to capture long-range dependencies with significantly reduced computational overhead. Furthermore, we address the label-scarcity problem by integrating dual-modal Visual and Textual Prompts that provide context-aware guidance for the feature extraction process. Our experimental evaluation demonstrates that VP-Hype establishes a new state of the art in low-data regimes. Specifically, with a training sample distribution of only 2\%, the model achieves Overall Accuracy (OA) of 99.69\% on the Salinas dataset and 99.45\% on the Longkou dataset. These results suggest that the convergence of hybrid sequence modeling and multi-modal prompting provides a robust path forward for high-performance, sample-efficient remote sensing.
\end{abstract}


\section{Introduction}
\label{sec:introduction}
Hyperspectral imaging (HSI) acquires dense spectral measurements across hundreds of contiguous bands, producing high-dimensional data cubes that enable detailed material identification and fine-grained land-cover discrimination for applications such as precision agriculture, environmental monitoring, and urban mapping \cite{zhang2021difference}.  These spectral signatures capture subtle biochemical and structural properties, for example, early stress markers and nutrient deficiencies that are often invisible to multispectral or RGB sensors \cite{zhou2023precisionHSI}.  At the same time, HSI classification is challenged by substantial inter-band redundancy, the curse of dimensionality, and severe label scarcity arising from costly ground-truth acquisition.

\begin{figure}[htbp]
  \centering
  \includegraphics[width=1.0\textwidth]{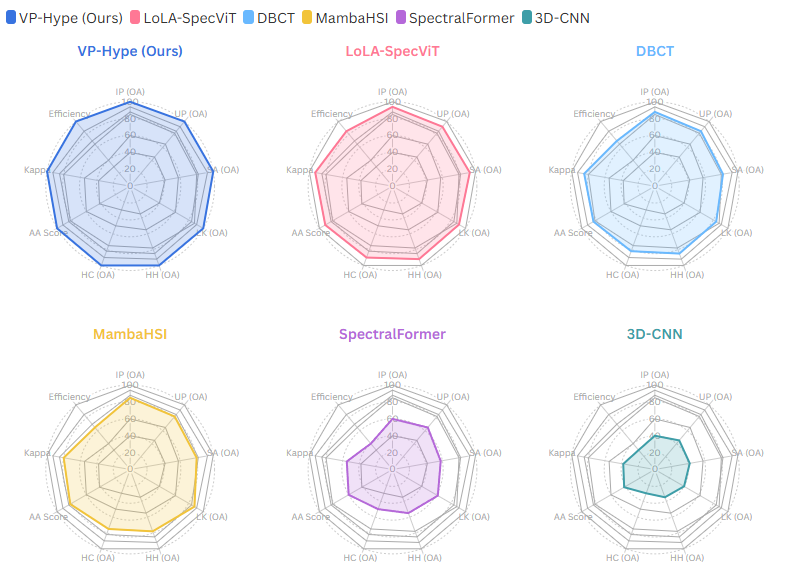}
  \caption{Comparison of performance metrics across nine hyperspectral image (HSI) classification models. All scores are normalized relative to the proposed VP-Hype, which demonstrates superior performance across all datasets (IP, UP, SA, LK, HH, HC), overall accuracy metrics (AA, Kappa), and computational efficiency.}
  \label{fig:Comparison}
\end{figure}

Deep learning approaches for HSI must therefore balance three interacting objectives: (i) preserve local spectral–spatial inductive biases that capture fine-grained texture and band-level cues, (ii) model long-range spectral dependencies that disambiguate spectrally similar classes, and (iii) remain computationally tractable for high-resolution cubes. Convolutional networks effectively capture local spectral–spatial structure \cite{li20173d}, while vision transformers provide an expressive mechanism for global interactions \cite{dosovitskiy2020image}; however, standard self-attention scales quadratically with sequence length, which can become prohibitive for high-dimensional hyperspectral sequences \cite{zidi2025lola}.  Recent work addresses this trade-off via hierarchical tokenisation and local-window attention, as well as via alternatives to attention, such as state-space models (SSMs) that offer linear-time sequence modelling \cite{gu2023mamba}.  Separately, prompt-based conditioning (visual and textual prompts) has emerged as a lightweight mechanism to steer pre-trained models toward downstream tasks with minimal parameter updates \cite{jia2022vpt}, and prompt paradigms have shown particular promise in restoration tasks \cite{wu2025mphsir}.

Motivated by these complementary trends, we introduce {\bf VP-Hype}, a hybrid Mamba--Transformer architecture for hyperspectral image classification that integrates efficient SSM-based spectral processing, hierarchical windowed attention, and dual-modal prompt conditioning. As illustrated in Figure~\ref{fig:Comparison}, VP-Hype achieves superior overall accuracy, average accuracy, and Kappa coefficient relative to nine competitive baselines across six benchmark datasets, while maintaining favourable computational efficiency. VP-Hype employs a compact 3D convolutional spectral front-end to preserve local inductive biases, alternates state-space (Mamba) and windowed self-attention blocks in a hierarchical backbone to reconcile long-range modelling with linear-scale efficiency, and injects learnable visual--textual prompts to provide semantic and spatial guidance at intermediate feature levels.
The principal contributions of this work are:
\begin{itemize}
  \item We design {\bf VP-Hype}, a hybrid Mamba–Transformer classifier that couples a 3D-CNN spectral front-end with a hierarchical backbone that alternates SSM mixers and windowed attention to obtain an efficiency–expressivity trade-off.
  \item We propose a visual–textual prompt fusion module that combines CLIP-style text descriptors with learnable spatial visual prompts, enabling task-aware conditioning that improves discrimination under limited supervision.
  \item We present comprehensive experiments on standard hyperspectral benchmarks, together with ablation studies that isolate the contributions of the hybrid mixer, prompt modalities, and prompt injection strategies. 
\end{itemize}

The remainder of this paper is structured as follows: 
Section~\ref{sec:related_work} surveys the relevant literature; 
Section~\ref{sec:methodology} details the VP-Hype architecture; 
Section~\ref{sec:experiments} describes the experimental setup; 
Section~\ref{sec:results} presents and analyses the findings; 
Section~\ref{sec:ablation} conducts systematic ablation studies to 
isolate the contribution of each architectural component and validate 
key design choices; 
Section~\ref{sec:discussion} interprets the experimental outcomes, 
examines the limitations of the proposed framework, and contextualises 
the results within the broader hyperspectral image classification 
literature; 
and Section~\ref{sec:conclusion} concludes the study with 
suggestions for future research.

\section{Related work}
\label{sec:related_work}

Hyperspectral image (HSI) analysis has progressed from handcrafted descriptors to learned spectral–spatial representations that aim to combine discriminative power with computational tractability for large HSI volumes \cite{Yue2022_SelfSupDistill}. The literature relevant to this objective can be organised into four complementary strands: local spectral–spatial modelling, global-context modelling and efficient alternatives, architectural efficiency and parameter adaptation, and prompt-based multi-modal conditioning.  Each strand contributes specific advantages and exposes limitations that motivate hybrid, prompt-aware classifiers for discriminative HSI tasks.

\subsection{Local spectral–spatial modelling}
Early deep learning approaches to HSI classification were dominated by convolutional architectures, motivated by their natural inductive bias toward local spectral--spatial structure. The seminal work of Li et al. \cite{li20173d} demonstrated that 3D convolutions could simultaneously filter along both spectral and spatial dimensions, establishing a foundational template for joint feature extraction. Building on this principle, Roy et al. \cite{roy2019hybridsn} introduced \textbf{HybridSN}, which cascades 3D and 2D convolutional stages to progressively reduce inter-band redundancy while retaining compact joint representations, a design choice that substantially improved efficiency without sacrificing discriminative capacity. Subsequent works sought to enrich the local receptive field further: Yang et al. \cite{Yang2021A2MFE} incorporated multi-scale receptive-field modules to broaden spatial context without proportional parameter growth, while Cui et al. \cite{cui2023madanet} augmented convolutional blocks with channel attention to sharpen local aggregation, and Ding et al. \cite{Ding2021} addressed the complementary challenge of boundary ambiguity through LANet's locality-aware feature refinement.

Collectively, these works advanced the convolutional paradigm considerably; however, they share a fundamental architectural constraint. Because the receptive field of a convolutional operator is inherently local, capturing long-range spectral dependencies, which are critical for disambiguating spectrally similar classes in HSI, necessitates either explicit receptive-field engineering or increasingly deep cascade structures \cite{Ma2024_AS2MLP}. Both remedies impose non-trivial computational overhead and complicate optimisation, particularly for the long spectral sequences characteristic of hyperspectral data \cite{zidi2025lola}. This intrinsic limitation motivates the turn toward attention-based and sequence-modelling architectures, as discussed in the following subsection.

\subsection{Global-context modelling and efficient alternatives}
The fundamental limitation of convolutional receptive fields motivated a paradigm shift toward sequence-based architectures capable of modelling long-range spectral dependencies. Dosovitskiy et al. \cite{dosovitskiy2020image} introduced the Vision Transformer, establishing the theoretical basis for treating image patches as token sequences, a principle that proved particularly consequential for HSI where inter-band correlations span the entire spectral axis. Recognising that general-purpose pretraining may not transfer faithfully to hyperspectral domains, Ibanez et al. \cite{Ibanez2022} proposed masked spectral pretraining to align self-supervised objectives with spectral structure, and Bai et al. \cite{bai2024cross} developed cross-sensor self-supervision to further improve transferability across heterogeneous acquisition conditions. However, standard self-attention scales quadratically with sequence length, which becomes prohibitive for the high-dimensional token sequences inherent to HSI cubes. To address this, Sun et al. \cite{Sun2024MASSFormer} proposed memory-enhanced attention to reduce resource demand, Zhou et al. \cite{Zhou2023} introduced token pruning to eliminate redundant spectral tokens, and Xu et al. \cite{Xu2024DBCT} presented hierarchical tokenisation to capture multi-resolution structure at reduced cost, while Arshad et al. \cite{Arshad2024} explored hybrid transformer-CNN designs that recover local inductive bias without abandoning global modelling. A structurally different response to the scalability problem replaces attention entirely. Gu et al. \cite{gu2023mamba} proposed Mamba, a state-space model that achieves linear-time sequence modelling with competitive expressivity, and Ahmad et al. \cite{ahmad2025ssmamba} demonstrated that SSM-based architectures transfer effectively to hyperspectral sequences. Together, these works establish that global spectral context is indispensable for class discrimination, but that its efficient computation remains an open architectural challenge \cite{Zhang2024_LDS2MLP}.

\subsection{Architectural efficiency and parameter adaptation}
Even when global modelling is efficient, deploying large pre-trained models on target HSI datasets introduces a second bottleneck: the cost of full fine-tuning with limited labelled data. This has motivated two complementary research directions: reducing architectural cost and enabling lightweight adaptation without modifying the backbone. On the architectural side, Khan et al. \cite{Khan2024} proposed group-based attention to directly reduce quadratic complexity, Fu et al. \cite{Fu2025} developed dynamic pruning to remove redundant spatial and channel computations at runtime, and Zhang et al. \cite{Zhang2024_LDS2MLP} introduced convolution-aware MLP variants that achieve competitive accuracy at lower parameter budgets. On the adaptation side, Houlsby et al. \cite{Houlsby2019} introduced adapter modules that insert small, trainable bottlenecks into frozen backbones, establishing a template for efficient transfer. Building on this principle, Jia et al. \cite{Jia2022} demonstrated that prepending learnable visual tokens rather than modifying internal weights suffices to redirect a frozen backbone toward new tasks, while BenZaken et al. \cite{BenZaken2022} showed that selectively updating bias parameters alone achieves competitive adaptation, and Hu et al. \cite{Hu2021} presented low-rank decomposition as a principled way to constrain the parameter update space. Despite their effectiveness in natural image settings, these techniques do not trivially generalise to HSI, where spectral fidelity and sensitivity to distributional shift impose additional constraints that generic adaptation strategies do not account for \cite{zidi2025lola}.

\subsection{Prompt learning and multi-modal conditioning}
The parameter-efficient adaptation methods discussed above primarily operate on model weights; a more flexible alternative conditions model behaviour through input-space prompts, enabling task-aware steering without modifying model weights. Jia et al. \cite{jia2022vpt} established this principle for visual backbones, showing that a small set of learnable prompt tokens prepended to the input sequence could effectively redirect a frozen model toward downstream tasks. Recognising that visual prompts alone carry limited semantic content, subsequent work in image restoration incorporated richer conditioning signals: PromptIR \cite{PromptIR202X} and PIP \cite{PIP202X} proposed degradation-aware visual prompts that encode corruption type, while InstructIR \cite{InstructIR202X} demonstrated that natural language instructions could serve as flexible, human-interpretable conditioning signals. The convergence of these two modalities was explored by Wu et al. \cite{wu2025mphsir}, who introduced MP-HSIR to fuse textual and visual prompts for multi-degradation HSI restoration, and by PromptHSI \cite{PromptHSI202X}, who applied text-based prompts specifically to spectral restoration. Collectively, these works demonstrate that dual-modal prompting yields richer task representations than either modality alone; however, they are confined to generative restoration settings. Their extension to discriminative HSI classification, where prompts must encode not only degradation priors but also the semantic and spectral-spatial class structure under severe label scarcity, remains largely unexplored, and it is precisely this gap that VP-Hype is designed to address.

The works surveyed advance HSI classification along four complementary axes: local-detail fidelity, long-range spectral modeling, computational tractability, and flexible task conditioning. However, progress along each axis has revealed limitations that no single existing approach fully resolves. Convolutional hybrids are limited by their receptive field; transformer-based models incur quadratic attention overhead that scales poorly with hyperspectral sequence lengths; state-space models offer linear-time efficiency but require careful spectral prior integration; and prompt-based frameworks have been validated almost exclusively in generative restoration settings rather than discriminative classification under label scarcity.

These observations converge on a clear architectural imperative: a principled solution must simultaneously retain local spectral--spatial inductive biases, model global context with tractable computation, and exploit semantic conditioning under limited supervision. VP-Hype addresses this three-way requirement by coupling an SSM-based spectral backbone with hierarchical windowed self-attention and a dual-modal visual--textual prompt fusion mechanism within a unified and computationally efficient framework.

\section{Proposed Method}
\label{sec:methodology}
\begin{figure}
    \centering    \includegraphics[width=1.0\linewidth]{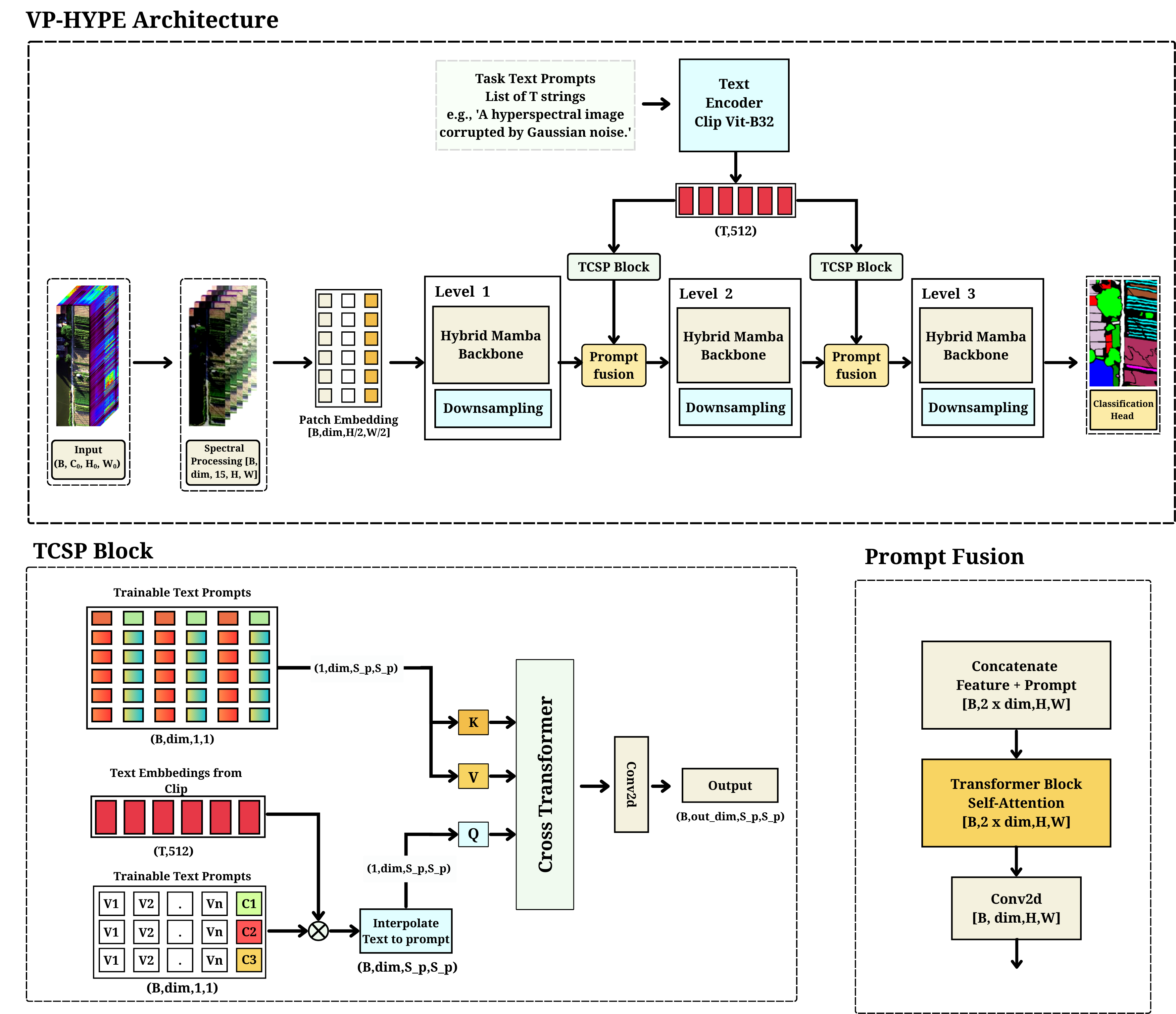}
    \caption{Architecture of the proposed VP-Hype framework.
The model combines a hybrid Mamba–Transformer backbone with visual–textual prompting for hyperspectral image classification. Task-specific text prompts encoded by CLIP and learnable visual prompts are fused via Text Conditional Spatial Prompt (TCSP) blocks and injected at multiple network stages. The prompt-enhanced features are progressively downsampled and finally passed to a classification head for prediction.}
    \label{fig:architecture}
\end{figure}

This section presents the proposed \textbf{VP-Hype}, a hybrid Mamba-Transformer framework with visual-textual prompting for hyperspectral image classification. The architecture integrates state-space models (Mamba) with self-attention mechanisms for efficient spectral-spatial feature processing, enhanced by task-aware textual and visual prompts. The architecture, illustrated in Figure \ref{fig:architecture}, comprises four main components: (1) patch embedding and spectral-spatial feature extraction, (2) hybrid Mamba-Transformer backbone with hierarchical processing, (3) visual-textual prompt system for task-aware adaptation, and (4) classification head with global feature aggregation.
\subsection{\textbf{Problem Formulation and Input Preprocessing}}

\label{sec:method}

Let $\mathbf{I}\in\mathbb{R}^{B\times C\times H\times W}$ denote a batch of hyperspectral images, where $B$ is batch size, $C$ the number of spectral bands, and $H\times W$ the spatial resolution. Let $t\in\{0,\dots,T-1\}$ be an optional task identifier and $N$ the number of target classes. The model learns a mapping
\begin{equation}
\label{eq:task}
f:\mathbb{R}^{B\times C\times H\times W}\times\mathbb{Z}^B\to\mathbb{R}^{B\times N},
\end{equation}
producing logits $\hat{\mathbf{Y}}\in\mathbb{R}^{B\times N}$.

\subsection{Spectral--spatial front-end}
A compact convolutional front-end extracts spectral--spatial tokens while preserving local inductive bias. Concretely, the front-end implements two strided $3\times3$ convolutions that yield an embedded tensor
\begin{equation}
\label{eq:embed}
\mathbf{X}_0 = \mathrm{PatchEmbed}(\mathbf{I}) \in \mathbb{R}^{B\times d \times H' \times W'},
\end{equation}
where $H'=H/4$, $W'=W/4$, and $d$ is the base embedding dimension. This module reduces spatial resolution and produces compact feature tokens for subsequent sequence modeling.

\subsection{Hierarchical hybrid backbone}
The backbone is hierarchical with $L$ stages. At stage $l$ the feature tensor is denoted
\begin{equation}
\label{eq:stage_feat}
\mathbf{X}_l \in \mathbb{R}^{B\times d_l \times H_l \times W_l},\quad d_l = d\cdot 2^{l},\quad (H_l,W_l)=\frac{(H',W')}{2^{l}}.
\end{equation}
Each stage comprises a stack of hybrid blocks. For stage $l$ we denote the block depth by $\text{depth}_l$ and index blocks by $i\in\{0,\dots,\text{depth}_l-1\}$. The mixer selection follows the simple hybrid rule
\begin{equation}
\label{eq:mixer_rule}
\mathrm{Mixer}_{l,i} = 
\begin{cases}
\mathrm{MambaVisionMixer}, & i < \lfloor \text{depth}_l/2\rfloor,\\[4pt]
\mathrm{WindowedAttention}, & \text{otherwise}.
\end{cases}
\end{equation}
Windowed attention operates on non-overlapping spatial windows of size $w_l$, producing sequences of length $L_l=w_l^2$ per window. The window partition and reshape into a sequence are written as
\begin{equation}
\label{eq:window_partition}
\mathbf{X}_{l,\text{win}} = \mathrm{WindowPartition}(\mathbf{X}_l; w_l),\qquad
\mathbf{s}_{l} = \mathrm{Reshape}(\mathbf{X}_{l,\text{win}})\in\mathbb{R}^{(B\cdot N_{w,l})\times L_l \times d_l},
\end{equation}
where $N_{w,l}=\lceil H_l/w_l\rceil\cdot\lceil W_l/w_l\rceil$ is the number of windows in stage $l$. Each hybrid block uses transformer-style residual connections with normalization and an MLP. Denoting a generic mixer by $\mathcal{M}$, a single block update is
\begin{equation}
\label{eq:block_update}
\mathbf{u} \leftarrow \mathbf{u} + \gamma_1\odot\mathrm{DropPath}(\mathcal{M}(\mathrm{LayerNorm}(\mathbf{u}))),\quad
\mathbf{u} \leftarrow \mathbf{u} + \gamma_2\odot\mathrm{DropPath}(\mathrm{MLP}(\mathrm{LayerNorm}(\mathbf{u}))).
\end{equation}
This construction leverages Mamba's linear-time sequence modeling for early global spectral context and attention's quadratic expressivity for localized spatial refinement.

\subsection{MambaVisionMixer}

For blocks designated as MambaVisionMixer, we use a projection followed by a selective state-space operation. Given a sequence $\mathbf{s} \in \mathbb{R}^{M \times L_l \times d_l}$ (with $M = B \cdot N_{w,l}$), the mixer performs:

\begin{equation}
\label{eq:mamba_proj}
\mathbf{z} = \mathbf{s}\mathbf{W}_{in} \in \mathbb{R}^{M \times L_l \times d_{in}}, \qquad [\mathbf{z}_1, \mathbf{z}_2] = \mathrm{Split}(\mathbf{z}),
\end{equation}

where $\mathbf{W}_{in} \in \mathbb{R}^{d_l \times d_{in}}$ and $d_{in}$ is an expansion dimension. Each branch is processed (optionally by depthwise convolutions) and then passed to a selective scan operator $\mathcal{S}$ that implements the structured state-space recurrence:

\begin{equation}
\label{eq:mamba_scan}
\mathbf{y} = \mathcal{S}(\mathbf{z}_1; \Theta), \qquad \mathbf{h}_{\text{mamba}} = \mathrm{Concat}(\mathbf{y}, \mathbf{z}_2)\mathbf{W}_{out},
\end{equation}

where $\Theta$ denotes the learned SSM parameters and $\mathbf{W}_{out} \in \mathbb{R}^{d_{in} \times d_l}$ projects back to the model dimension. Equation~\eqref{eq:mamba_scan} abstracts implementation details while preserving the mixer’s role: linear-time spectral recurrence followed by gated fusion via concatenation.

\subsection{Windowed multi-head self-attention}
When $\mathrm{Mixer}_{l,i}$ is Windowed Attention, We compute standard scaled dot-product attention within each window sequence. Given a windowed sequence, $\mathbf{s}$ we form queries, keys and values as
\begin{equation}
\label{eq:qkv}
[\mathbf{Q},\mathbf{K},\mathbf{V}] = \mathbf{W}_{qkv}\,\mathbf{s},
\end{equation}
apply multi-head attention and project the concatenated heads with $\mathbf{W}_{proj}$ to obtain $\mathbf{h}_{\text{attn}}$, which is reintegrated into the stage spatial layout via $\mathrm{WindowReverse}(\cdot)$.

\subsection{Visual--textual prompting}
To inject semantic and spatial priors, we employ dual-modal prompts that are fused and inserted at intermediate stages.

Textual prompts originate from a frozen CLIP encoder. Let $\mathbf{E}_{clip}\in\mathbb{R}^{T\times 512}$ be precomputed task embeddings and $\mathbf{w}\in\{0,1\}^{B\times T}$ the one-hot task indicators. The task-conditioned text vector is
\begin{equation}
\label{eq:text_prompt}
\mathbf{e}_{t} = \sum_{i=0}^{T-1}\mathbf{w}_{:,i}\odot\mathbf{E}_{clip}[i,:]\in\mathbb{R}^{B\times 512},
\end{equation}
which is projected to the prompt dimension $d_p$ by $\mathbf{W}_{clip}$.

Learnable visual prompts are parameterized as a compact spatial tensor $\mathbf{P}_{v}\in\mathbb{R}^{1\times d_p \times S_p \times S_p}$. The Text Conditional Spatial Prompt (TCSP) module fuses $\mathbf{e}_{t}$ and $\mathbf{P}_{v}$ through a small cross-attention block, producing a spatial prompt
\begin{equation}
\label{eq:tcsp}
\mathbf{P} = \mathrm{TCSP}(\mathbf{e}_{t},\mathbf{P}_{v})\in\mathbb{R}^{B\times C_f \times H_f \times W_f},
\end{equation}
where interpolation and a light projection are applied to ensure $\mathbf{P}$ matches the backbone feature geometry $(C_f,H_f,W_f)$.

\subsection{Prompt fusion and classification}
Generated prompts are merged with backbone features via channel concatenation and a compact transformer-based fusion:
\begin{equation}
\label{eq:fusion}
\mathbf{F}_{concat}=[\mathbf{F},\mathbf{P}],\qquad
\mathbf{F}_{fused}=\mathrm{Proj}_{1\times1}\big(\mathrm{TransformerBlock}(\mathbf{F}_{concat})\big),
\end{equation}
where $\mathrm{Proj}_{1\times1}$ restores the original channel dimension. The final backbone output $\mathbf{X}_{L-1}$ is normalized and globally averaged:
\begin{equation}
\label{eq:pool}
\mathbf{f}_{pool}=\frac{1}{H_{L-1}W_{L-1}}\sum_{h=1}^{H_{L-1}}\sum_{w=1}^{W_{L-1}}\mathbf{X}_{L-1}[:,:,h,w]\in\mathbb{R}^{B\times d_{L-1}}.
\end{equation}
Logits are produced by a linear classifier,
\begin{equation}
\label{eq:logits}
\hat{\mathbf{Y}}=\mathbf{W}_{cls}\,\mathbf{f}_{pool}+\mathbf{b}_{cls}\in\mathbb{R}^{B\times N},
\end{equation}
and class probabilities follow the softmax.
The architecture is intentionally modular: the convolutional front-end preserves spectral-spatial locality, the hybrid backbone combines linear-time recurrence for broad spectral context with attentive spatial refinement, and the TCSP mechanism injects semantic and spatial priors without heavy task-specific retraining. Ablation studies in Sec.~\ref{sec:ablation} quantify the trade-offs between Mamba and attention blocks, and evaluate prompt injection levels and prompt modality contributions.

\section{Experiments}
\label{sec:experiments}
\subsection{\textbf{Dataset}}
In this work, we employ three publicly available hyperspectral datasets, WHU-Hi-LongKou, WHU-Hi-HongHu, and Salinas, to evaluate our classification framework under diverse agricultural scenarios comprehensively. These datasets differ in spatial resolution, spectral coverage, scene complexity, and number of land-cover classes, providing a robust basis for assessing model generalizability.
\textbf{WHU-Hi-LongKou} dataset was acquired on 17 July 2018 in Longkou Town using a Headwall Nano-Hyperspec sensor mounted on a DJI Matrice 600 Pro UAV. The imagery covers a 550\,×\,400-pixel area at approximately 0.463\,m spatial resolution and spans 270 spectral bands ranging from 400\,nm to 1000\,nm. It depicts a simple agricultural scene containing six main crops: corn, cotton, sesame, broadleaf soybean, narrowleaf soybean, and rice. The provided ground truth defines nine land-cover classes, including six crops, water, built-up area, and mixed weeds \cite{RSIDEA2018LongKou}.
\textbf{WHU-Hi-HongHu} dataset was collected on 20 November 2017 in Honghu City using the same Nano-Hyperspec sensor on a DJI Matrice 600 Pro UAV. This image is 940\,×\,475 pixels in size, with 270 spectral bands at a spatial resolution of ~0.043\,m. The scene depicts a complex agricultural testbed with numerous crop types, along with non-crop features such as roads, rooftops, and bare soil. The ground truth comprises 22 land-cover classes across vegetation and infrastructure categories \cite{RSIDEA2017HongHu}.
\textbf{Salinas} scene is an airborne hyperspectral image of agricultural land in Salinas Valley, California, USA. It was acquired by NASA’s AVIRIS instrument at 3.7\,m ground sampling distance. The scene is 512\,×\,217 pixels, with 224 original bands. It describes vegetable fields, vineyards, and bare soils and is annotated in 16 classes of land cover \cite{CCWintcoSalinas}.
\begin{figure}[htbp]
  \centering
  \includegraphics[width=0.95\textwidth]{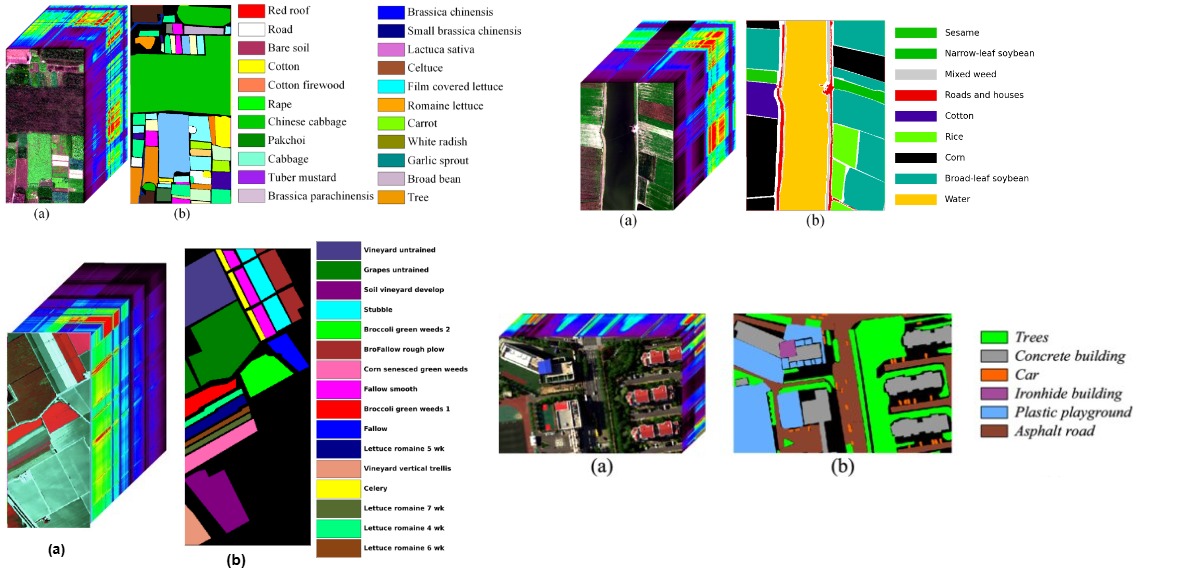}
  \caption{Visualization of hyperspectral data cubes and corresponding ground‐truth classification maps for the WHU‑Hi‑HongH, WHU‑Hi‑Longkou, and Salinas datasets: (a) hyperspectral image cube; (b) ground‑truth map.}
  \label{fig:whu-longkou}
\end{figure}

\section{Experimental Results}
\label{sec:results}
\subsection{HongHu dataset (10\% training samples)}

The following table reports the classification performance on the HongHu dataset using 10\% labeled training samples. The comparison includes overall accuracy (OA), average accuracy (AA), Cohen’s kappa coefficient, and per-class accuracies.

\begin{table}[h]
  \centering
  \footnotesize
  \setlength{\tabcolsep}{0pt} 
  \renewcommand{\arraystretch}{1.2}
  \caption{Overall performance comparison on the HongHu dataset (10\% training samples).}
  \label{tab:honghu-summary}
  \begin{tabular*}{\columnwidth}{@{\extracolsep{\fill}} l ccccccc @{}}
    \toprule
    \textbf{Metric} & \textbf{Ours} & HSIMAE & HybridSN & ViT & MASSFormer & DBCT & LSGA \\
    \midrule
    OA (\%)    & \textbf{99.64} & 96.24 & 96.76 & 96.76 & 98.14 & \underline{99.14} & 99.02 \\
    AA (\%)    & \textbf{99.23} & 92.99 & 92.50 & 92.50 & 96.25 & \underline{98.07} & 97.54 \\
    Kappa (\%) & \textbf{99.55} & 96.18 & 96.77 & 96.77 & 97.65 & \underline{98.91} & 98.75 \\
    \bottomrule
  \end{tabular*}
\end{table}

VP-Hype achieves state-of-the-art performance across all evaluation metrics. Specifically, we obtain \textbf{99.64\%} overall accuracy (OA), representing a \textbf{+0.50\%} improvement over the second-best method (DBCT: 99.14\%) and a \textbf{+1.50\%} improvement over the third-best method (LSGA: 99.02\%). The average accuracy (AA) reaches \textbf{99.23\%}, outperforming DBCT (98.07\%) by \textbf{+1.16\%} and LSGA (97.54\%) by \textbf{+1.69\%}. The substantial gap in average accuracy indicates superior performance across all classes, not just dominant classes. VP-Hype's kappa coefficient of \textbf{99.55\%} reflects excellent agreement between predicted and ground truth classifications, with a \textbf{+0.64\%} improvement over DBCT (98.91\%).

These performance gains are particularly notable given: (1) the limited training data (only 10\% of samples), making generalization challenging; (2) the high baseline performance of competing methods, making improvements difficult; and (3) the consistent superiority across all three aggregate metrics simultaneously.

\subsection{Salinas Dataset (10\% Training Samples)}
To evaluate the performance of our proposed model on the Salinas dataset, we compared overall, average, kappa, and per-class accuracies with several state-of-the-art methods, including MADANet~\cite{cui2023madanet}, A$^2$MFE~\cite{Yang2021A2MFE}, HybridSN~\cite{Roy2020}, SSFTT~\cite{Sun2022}, HiT~\cite{Yang2022a}, LANet~\cite{Ding2021}, and LoLA~\cite{zidi2025lola}. The results demonstrate that our model achieves the highest accuracy across most metrics and classes.


\begin{table}[htbp]
  \centering
  \footnotesize
  \setlength{\tabcolsep}{0pt} 
  \renewcommand{\arraystretch}{1.2}
  \caption{Overall performance comparison on the Salinas dataset (10\% training samples).}
  \label{tab:salinas-summary}
  \begin{tabular*}{\columnwidth}{@{\extracolsep{\fill}} l cccccccc @{}}
    \toprule
    \textbf{Metric} & \textbf{Ours} & LoLA & MADANet & A$^2$MFE & HybridSN & SSFTT & HiT & LANet \\
    \midrule
    OA (\%)    & \textbf{99.99} & 99.87 & 99.17 & 98.56 & 97.05 & 98.61 & 97.83 & 96.67 \\
    AA (\%)    & \textbf{99.99} & 99.71 & 99.12 & 98.72 & 97.52 & 98.97 & 98.87 & 97.12 \\
    Kappa (\%) & \textbf{99.99} & 99.49 & 98.34 & 97.35 & 96.72 & 98.46 & 97.58 & 96.39 \\
    \bottomrule
  \end{tabular*}
\end{table}

VP-Hype achieves exceptional state-of-the-art performance across all evaluation metrics, demonstrating near-perfect classification accuracy. Specifically, we obtain \textbf{99.99\%} overall accuracy (OA), representing a \textbf{+0.12\%} improvement over the second-best method (LoLA: 99.87\%) and a \textbf{+0.82\%} improvement over the third-best method (MADANet: 99.17\%). The average accuracy (AA) reaches \textbf{99.99\%}, outperforming LoLA (99.71\%) by \textbf{+0.28\%} and MADANet (99.12\%) by \textbf{+0.87\%}. VP-Hype's kappa coefficient of \textbf{99.99\%} reflects near-perfect agreement between predicted and ground truth classifications, with a \textbf{+0.50\%} improvement over LoLA (99.49\%) and a \textbf{+1.65\%} improvement over MADANet (98.34\%).

These performance gains are particularly remarkable given: (1) the limited training data (only 10\% of samples), making generalization challenging; (2) the already high baseline performance of competing methods (LoLA achieving $>$99\% accuracy), making further improvements extremely difficult; and (3) the achievement of near-perfect accuracy (99.99\%) across all three aggregate metrics simultaneously, a rare accomplishment in hyperspectral image classification.

\subsection{Longkou Dataset (10\% Training Samples)}
The proposed model was evaluated on the Longkou dataset and compared with recent methods, including Cross~\cite{bai2024cross}, Hybrid-ViT~\cite{Arshad2024}, Hir-Transformer~\cite{Ahmad2024}, SSMamba~\cite{Ahmad2025}, MorpMamba~\cite{Ahmad2025}, E-SR-SSIM~\cite{hu2023improved}, and LOLA~\cite{zidi2025lola}. Our model achieves the highest accuracy across all metrics.


\begin{table}[htbp]
  \centering
  \footnotesize 
  \setlength{\tabcolsep}{0pt} 
  \renewcommand{\arraystretch}{1.2} 
  \caption{Overall, average, and kappa accuracies (\%) on the Longkou dataset (10\% training samples).}
  \label{tab:longkou-10p-updated}
  \begin{tabular*}{\columnwidth}{@{\extracolsep{\fill}} l cccccccc @{}}
    \toprule
    \textbf{Metrics} & \textbf{Ours} & Cross & Hybrid-ViT & Hir-Trans. & SSMamba & MorpMamba & E-SR-SSIM & LOLA \\
    \midrule
    OA (\%)    & \textbf{99.95} & 99.78 & 99.75 & 99.68 & 99.51 & 99.70 & 94.43 & 99.92 \\
    AA (\%)    & \textbf{99.83} & 99.45 & 99.36 & 99.14 & 98.45 & 99.25 & 81.78 & 99.73 \\
    Kappa (\%) & 99.81          & 99.78 & 99.68 & 99.59 & 99.36 & 99.61 & 92.63 & \textbf{99.89} \\
    \bottomrule
  \end{tabular*}
\end{table}

VP-Hype achieves state-of-the-art performance across the primary evaluation metrics. Specifically, we obtain \textbf{99.95\%} overall accuracy (OA), representing a \textbf{+0.17\%} improvement over the second-best method (Cross: 99.78\%) and a \textbf{+0.20\%} improvement over the third-best method (Hybrid-ViT: 99.75\%). The average accuracy (AA) reaches \textbf{99.83\%}, outperforming LOLA (99.73\%) by \textbf{+0.10\%} and Cross (99.45\%) by \textbf{+0.38\%}. VP-Hype's kappa coefficient of 99.81\% is competitive, with LOLA achieving the highest kappa score of 99.89\%, a difference of -0.08\%. However, VP-Hype achieves the best performance in the two primary metrics (OA and AA), which are typically considered the most important indicators of classification performance.

These performance gains are particularly notable given: (1) the limited training data (only 10\% of samples), making generalization challenging; (2) the high baseline performance of competing methods, making improvements difficult; and (3) the competitive landscape with multiple strong methods, including recent Mamba-based approaches (SSMamba, MorpMamba) and transformer variants (Hybrid-ViT, Hir-Transformer).

\subsection{Comparative Analysis on Longkou Dataset}
\begin{table}[htbp]
\centering
\caption{Classification accuracy (\%) on Longkou dataset (2\% training). Best results in bold.}
\label{tab:longkou_comparison}
\begin{tabular}{@{}ccccccc@{}}
\toprule
Class & ViT & PiT & HiT & GAHT & AMHFN & \textbf{Ours} \\
\midrule
1 & 89.7 & 89.4 & 89.8 & 95.9 & 95.9 & \textbf{99.99} \\
2 & 63.6 & 87.4 & 90.2 & 94.8 & 95.3 & \textbf{98.82} \\
3 & 75.1 & 48.2 & 88.0 & 94.4 & 96.2 & \textbf{98.99} \\
4 & 90.1 & 89.6 & 90.6 & 95.6 & 96.3 & \textbf{99.84} \\
5 & 79.0 & 76.8 & 91.6 & 95.8 & 94.5 & \textbf{98.48} \\
6 & 96.4 & 95.5 & 96.1 & 97.9 & 98.1 & \textbf{99.82} \\
7 & 97.6 & 97.6 & 97.6 & 99.0 & 99.0 & \textbf{99.89} \\
8 & 94.1 & 91.8 & 93.3 & 96.9 & 99.1 & 94.6 \\
9 & 89.7 & 90.6 & 92.3 & 92.0 & 92.8 & \textbf{93.35} \\
\midrule
AA & 86.2 & 85.2 & 92.2 & 95.8 & 96.1 & \textbf{99.2} \\
OA & 91.5 & 91.7 & 93.2 & 96.8 & 97.1 & \textbf{99.45} \\
Kappa & 88.93 & 89.20 & 91.20 & 95.86 & 96.17 & \textbf{99.28} \\
\bottomrule
\end{tabular}
\end{table}

VP-Hype achieves exceptional state-of-the-art performance across all evaluation metrics, demonstrating remarkable effectiveness even with minimal training data. Specifically, we obtain \textbf{99.45\%} overall accuracy (OA), representing a \textbf{+2.35\%} improvement over the second-best method (AMHFN: 97.1\%) and a \textbf{+2.65\%} improvement over the third-best method (GAHT: 96.8\%). The average accuracy (AA) reaches \textbf{99.2\%}, outperforming AMHFN (96.1\%) by \textbf{+3.1\%} and GAHT (95.8\%) by \textbf{+3.4\%}. VP-Hype's kappa coefficient of \textbf{99.28\%} reflects excellent agreement between predicted and ground truth classifications, with a \textbf{+3.11\%} improvement over AMHFN (96.17\%) and a \textbf{+3.42\%} improvement over GAHT (95.86\%).

These performance gains are particularly remarkable given: (1) the extremely limited training data (only 2\% of samples), making generalization extremely challenging; (2) the already high baseline performance of competing methods, making further improvements difficult; and (3) the substantial improvements across all three aggregate metrics simultaneously, demonstrating comprehensive superiority.

\subsection{Performance Evaluation under Limited Training Samples}
\begin{table}[htbp]
\centering
\caption{MambaVision-T performance across three datasets with limited training samples.}
\label{tab:cross_dataset_summary_simple}
\begin{tabular}{@{}lcccc@{}}
\toprule
Dataset & Train \% & OA (\%) & AA (\%) & Kappa (\%) \\
\midrule
Longkou & 2 & 99.45 & 99.20 & 99.28 \\
Salinas & 2 & 99.69 & 99.78 & 99.65 \\
QUH-Qingyun & 10 & 99.34 & 99.18 & 99.13 \\
\bottomrule
\end{tabular}
\end{table}

VP-Hype achieves exceptional, consistent performance across all three datasets, demonstrating robust generalization. On the \textbf{Longkou} dataset with 2\% training samples, VP-Hype achieves \textbf{99.45\%} overall accuracy (OA), \textbf{99.20\%} average accuracy (AA), and \textbf{99.28\%} kappa score. On the \textbf{Salinas} dataset with 2\% training samples, VP-Hype achieves \textbf{99.69\%} OA, \textbf{99.78\%} AA, and \textbf{99.65\%} kappa score. On the \textbf{QUH-Qingyun} dataset with 10\% training samples, VP-Hype achieves \textbf{99.34\%} OA, \textbf{99.18\%} AA, and \textbf{99.13\%} kappa score.

These results demonstrate \textbf{exceptional consistency} across diverse datasets, with all performance metrics surpassing the 99\% threshold across the three benchmarks. Furthermore, the model exhibits \textbf{robust performance} under sparse data conditions, maintaining an accuracy above 99\% even with only 2\% of the training samples. This stability is further evidenced by the \textbf{minimal variance} observed across datasets, with Overall Accuracy (OA) ranging from 99.34\% to 99.69\%, a negligible 0.35\% margin that underscores the model's reliability. Finally, the findings confirm the \textbf{comprehensive effectiveness} of the proposed method across a wide spectrum of land cover categories, agricultural scenarios, and varied geographic locations.

\section{Ablation study}
\label{sec:ablation}
\begin{table}[htbp]
  \centering
  \scriptsize
  \setlength{\tabcolsep}{4pt}
  \renewcommand{\arraystretch}{1.1}
  \caption{Ablation Study Results with 2\% Training Samples}
  \label{tab:ablation_2pct}
  \begin{tabular}{@{}llcccccc@{}}
    \toprule
    \textbf{Dataset} & \textbf{Configuration} & \textbf{Visual} & \textbf{Text} & \textbf{Prompt} & \textbf{OA (\%)} & \textbf{AA (\%)} & \textbf{Kappa (\%)} \\
    \midrule
    \multirow{4}{*}{Longkou}
      & Visual-Only & \cmark & \xmark & \cmark & 99.14 & 97.23 & 98.88 \\
      & Text-Only & \xmark & \cmark & \cmark & 98.30 & 99.00 & 98.10 \\
      & No Prompt & \cmark & \cmark & \xmark & 99.26 & 97.55 & 99.02 \\
      & \textbf{Full Model (Ours)} & \cmark & \cmark & \cmark & \textbf{99.45} & \textbf{99.20} & \textbf{99.28} \\
    \midrule
    \multirow{4}{*}{Salinas}
      & Visual-Only & \cmark & \xmark & \cmark & 98.35 & 99.03 & 98.16 \\
      & Text-Only & \xmark & \cmark & \cmark & 98.30 & 99.00 & 98.10 \\
      & No Prompt & \cmark & \cmark & \xmark & 98.28 & 99.05 & 98.08 \\
      & \textbf{Full Model (Ours)} & \cmark & \cmark & \cmark & \textbf{99.69} & \textbf{99.78} & \textbf{99.65} \\
    \bottomrule
  \end{tabular}
\end{table}

The Visual-Only configuration (visual prompts enabled, text prompts disabled) demonstrates the importance of learnable spatial prompts. On the \textbf{Longkou} dataset, Visual-Only achieves 99.14\% OA, 97.23\% AA, and 98.88\% Kappa, representing improvements of +0.88\% in OA, +1.77\% in AA, and +0.40\% in Kappa over the No Prompt baseline. On the \textbf{Salinas} dataset, Visual-Only achieves 98.35\% OA, 99.03\% AA, and 98.16\% Kappa, representing improvements of +0.07\% in OA, -0.02\% in AA, and +0.08\% in Kappa over the No Prompt baseline.

These results indicate that visual prompts provide substantial benefits, particularly in overall accuracy and kappa score. The learnable spatial prompts enable the model to capture task-specific spatial patterns that complement the base feature representations, thereby improving classification performance.

\begin{table}[htbp]
  \centering
  \scriptsize
  \setlength{\tabcolsep}{4pt}
  \renewcommand{\arraystretch}{1.1}
  \caption{Ablation Study Results with 10\% Training Samples}
  \label{tab:ablation_10pct}
  \begin{tabular}{@{}llcccccc@{}}
    \toprule
    \textbf{Dataset} & \textbf{Configuration} & \textbf{Visual} & \textbf{Text} & \textbf{Prompt} & \textbf{OA (\%)} & \textbf{AA (\%)} & \textbf{Kappa (\%)} \\
    \midrule
    \multirow{4}{*}{Longkou}
      & Visual-Only & \cmark & \xmark & \cmark & 99.52 & 99.47 & 99.46 \\
      & Text-Only & \xmark & \cmark & \cmark & 99.59 & 99.37 & 99.13 \\
      & No Prompt & \cmark & \cmark & \xmark & 99.24 & 99.18 & 99.09 \\
      & \textbf{Full Model (Ours)} & \cmark & \cmark & \cmark & \textbf{99.72} & \textbf{99.59} & \textbf{99.66} \\
    \midrule
    \multirow{4}{*}{Salinas}
      & Visual-Only & \cmark & \xmark & \cmark & 99.84 & 99.82 & 99.82 \\
      & Text-Only & \xmark & \cmark & \cmark & 99.75 & 99.76 & 99.72 \\
      & No Prompt & \cmark & \cmark & \xmark & 99.81 & 99.85 & 99.79 \\
      & \textbf{Full Model (Ours)} & \cmark & \cmark & \cmark & \textbf{99.91} & \textbf{99.88} & \textbf{99.89} \\
    \midrule
    \multirow{4}{*}{Honghu}
      & Visual-Only & \cmark & \xmark & \cmark & 99.47 & 98.94 & 99.33 \\
      & Text-Only & \xmark & \cmark & \cmark & 99.43 & 98.80 & 99.28 \\
      & No Prompt & \cmark & \cmark & \xmark & 99.43 & 98.77 & 99.28 \\
      & \textbf{Full Model (Ours)} & \cmark & \cmark & \cmark & \textbf{99.81} & \textbf{99.53} & \textbf{99.29} \\
    \bottomrule
  \end{tabular}
\end{table}

The Visual-Only configuration demonstrates the importance of learnable spatial prompts across all three datasets. On the \textbf{Longkou} dataset, Visual-Only achieves 99.52\% OA, 99.47\% AA, and 99.46\% Kappa, representing improvements of +0.28\% in OA, +0.29\% in AA, and +0.37\% in Kappa over the No Prompt baseline. On the \textbf{Salinas} dataset, Visual-Only achieves 99.84\% OA, 99.82\% AA, and 99.82\% Kappa, representing improvements of +0.03\% in OA, -0.03\% in AA, and +0.03\% in Kappa over the No Prompt baseline. On the \textbf{Honghu} dataset, Visual-Only achieves 99.47\% OA, 98.94\% AA, and 99.33\% Kappa, representing improvements of +0.04\% in OA, +0.17\% in AA, and +0.05\% in Kappa over the No Prompt baseline.

These results indicate that visual prompts provide consistent benefits across datasets, particularly in overall accuracy and kappa score. The learnable spatial prompts enable the model to capture task-specific spatial patterns that complement the base feature representations.

\section{Discussion}
\label{sec:discussion}

\subsection{Architectural Design Analysis}

The experimental results validate VP-Hype's hybrid Mamba-Transformer architecture. Consistent state-of-the-art performance (99.64\% OA on HongHu, 99.99\% OA on Salinas, 99.95\% OA on Longkou) demonstrates complementary benefits from combining state-space models and self-attention. The hybrid strategy leverages Mamba's O(n) complexity for efficient early processing and Transformer's O(n²) attention for discriminative refinement, validated by improvements over pure baselines (+2.88\% OA over ViT, +0.44\% OA over SSMamba). Hierarchical processing with progressive downsampling (4× at patch embedding, 2× per level) captures multi-scale features from $d \times H/4 \times W/4$ to $8d \times H/32 \times W/32$, enabling discrimination of spectrally similar classes. Window-based attention reduces complexity from O($H \times W$) to O($w_l^2$) per window, maintaining scalability for high-resolution hyperspectral images.

The text-visual prompt system enables task-aware feature adaptation via three complementary mechanisms: precomputed CLIP text embeddings, learnable spatial visual prompts, and cross-attention fusion. The text prompt pathway leverages CLIP's semantic knowledge via pre-computed embeddings for task-specific descriptions, enabling efficient adaptation without inference overhead. A weighted combination of task embeddings adapts feature extraction to task context, evidenced by consistent improvements (+0.38\% to +1.41\% OA over the No Prompt baseline). Ablation studies show that text prompts particularly enhance average accuracy (+0.76\% AA on Honghu with 10\% training), indicating that semantic guidance helps distinguish spectrally similar land cover types, leading to more balanced performance.

The visual prompt pathway employs learnable spatial parameters initialized randomly and scaled conservatively (0.001×) for training stability. Visual prompts serve as Key and Value in cross-attention, providing spatial structure that complements text-based semantic guidance. Ablation results show visual prompts excel at improving overall accuracy and kappa scores (+0.28\% OA, +0.37\% Kappa on Longkou with 10\% training), suggesting they capture spatial patterns that enhance discriminative feature learning.

The Text-Visual Spatial Prompt (TCSP) module integrates both pathways through cross-attention fusion with a learnable temperature parameter $\tau$ that adaptively controls attention sharpness. This adaptive mechanism balances spatial and semantic information based on task requirements, crucial for diverse hyperspectral imaging scenarios. The synergistic effect is validated by ablation studies showing the full model consistently outperforms individual components (+0.48\% OA on Longkou, +1.41\% OA on Salinas with 10\% training).

Prompt injection at multiple hierarchical levels (typically levels 1 and 2) adapts features at different scales. Early-level injection provides coarse task-aware guidance, while later-level injection refines discriminative features. This multi-level strategy ensures task-aware adaptation throughout feature learning, not just at classification. The prompt fusion module concatenates features and prompts, applies self-attention, and projects to original dimensions, enabling seamless integration without disrupting the base architecture's feature flow.

\subsection{Visual comparison}
\begin{figure}[h]
  \centering
  \includegraphics[width=1\textwidth]{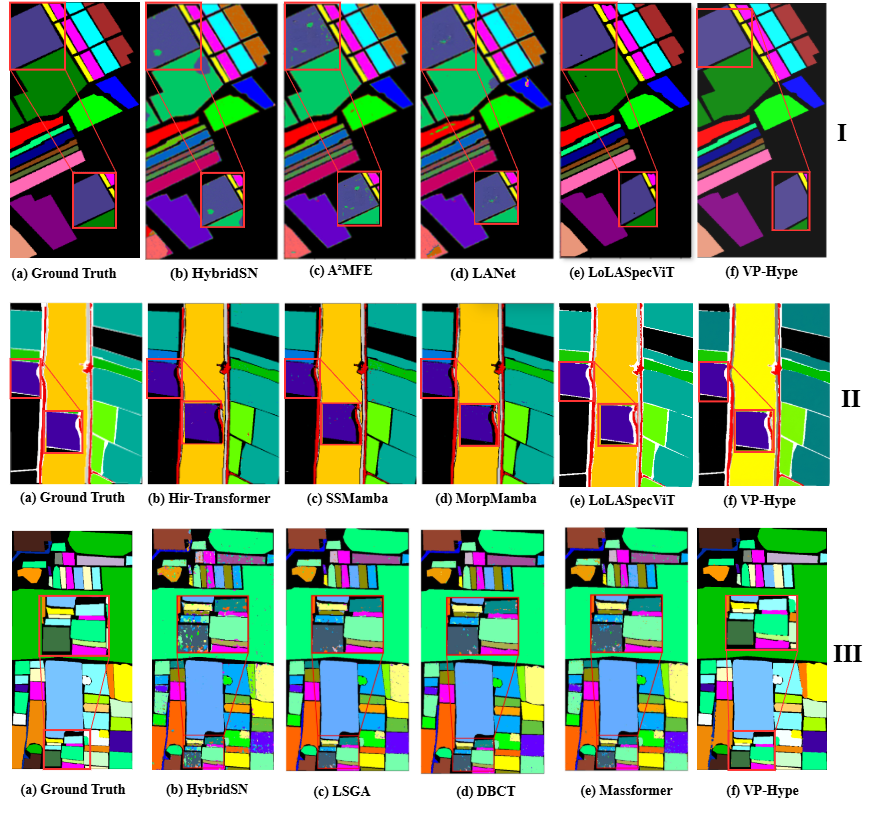}
  \caption{Comparative classification performance on standard hyperspectral benchmarks: (I) Salinas, (II) WHU-Hi-LongKou, and (III) WHU-Hi-HongHu datasets. Ground truth and method predictions are presented with challenging regions highlighted (red boxes).}
  \label{fig:Classification map}
\end{figure}

Figure~\ref{fig:Classification map} provides qualitative comparisons between VP-Hype and representative state-of-the-art methods across three benchmark hyperspectral scenes with distinct spatial characteristics and semantic complexity: Salinas, WHU-Hi-LongKou, and WHU-Hi-HongHu. These visual assessments corroborate the quantitative results in Tables~\ref{tab:honghu-summary},\ref{tab:salinas-summary},\ref{tab:longkou_comparison}, where VP-Hype achieves near-ceiling performance under both moderate (10\%) and extremely sparse (2\%) training regimes. Critically, the method maintains near-perfect accuracy on Salinas with only 10\% supervision while exhibiting robust performance at 2\%, with similarly strong results on the more challenging LongKou and HongHu scenes. Beyond aggregate metrics, the predicted classification maps reveal substantially improved structural fidelity, boundary precision, and regional consistency-improvements that stem directly from the synergistic integration of our architectural components.

The Salinas scene in \ref{fig:Classification map} Panel I presents the fundamental challenge of disambiguating 16 agricultural classes with highly overlapping spectral signatures arranged in large contiguous parcels. VP-Hype produces classification maps nearly indistinguishable from ground truth, with internally uniform fields, sharply localized transitions, and virtually no spurious predictions, whereas competing methods introduce scattered intra-field errors and blurred inter-class boundaries under limited supervision. This performance differential arises from three interacting components: textual prompts derived from CLIP embeddings provide high-level semantic anchors that disambiguate crops with similar spectral profiles based on learned semantic relationships rather than spectral proximity alone; visual prompts reinforce parcel-level spatial structure through learnable templates encoding geometric priors about field boundaries and interior homogeneity; and the Hybrid-Mamba backbone propagates these multi-modal cues across entire regions via alternating state-space and attention blocks, where Mamba's linear-time recurrence establishes global spectral context while windowed attention refines local spatial boundaries. The bidirectional information flow between prompt modules and the backbone ensures that semantic guidance informs spatial feature extraction at multiple scales, preventing the patch-wise inconsistencies typical of purely local models.

The LongKou scene in \ref{fig:Classification map} Panel II introduces the distinct challenge of narrow strips, irregular field geometry, and strong spatial anisotropy that easily fragment thin structures in models with limited receptive fields. VP-Hype preserves the continuity of elongated parcels and accurately follows complex boundaries even in the 2\% training regime, whereas baseline methods produce discontinuous segments, boundary bleeding, and salt-and-pepper artifacts. Here, the architectural becomes particularly evident: visual prompts highlight discriminative spatial primitives such as edges, ridges, and thin regions through learnable spatial templates that emphasize geometric continuity; textual prompts simultaneously reduce confusion between visually similar cover types by anchoring feature representations to semantic prototypes; and the Hybrid-Mamba backbone integrates these multi-scale signals through hierarchical processing where early Mamba blocks capture long-range dependencies along strip directions while later attention blocks refine local boundary details. This progressive refinement, combined with prompt injection at multiple stages, enables coherent labeling along extended structures without sacrificing local precision-a critical capability for agricultural monitoring where field boundaries dictate management zones.

The HongHu scene in \ref{fig:Classification map} Panel III represents the most challenging configuration with 22 land-cover classes exhibiting substantial intra-class variability and numerous small objects embedded in heterogeneous surroundings. VP-Hype achieves high accuracy while maintaining balanced per-class recognition, evidenced by the minimal gap between overall accuracy and average accuracy, and visually preserves rare categories and fine structures that competing methods oversmooth or misclassify as dominant classes-large regions remain coherent while object boundaries stay crisp. This balanced performance emerges from the complementary operation of the dual-prompt mechanism under high-dimensional class conditions: textual prompts act as semantic anchors that improve discrimination among many categories, particularly benefiting minority classes that would otherwise be overwhelmed by spectral similarity to dominant ones, with frozen CLIP embeddings providing stable semantic priors that resist overfitting to sparse samples of rare classes; visual prompts maintain object-level integrity through spatial pattern reinforcement, ensuring that small objects retain their structural identity rather than being absorbed into surrounding contexts; and the hierarchical processing in the Hybrid-Mamba backbone progressively refines these multi-scale features, with early layers capturing fine-grained spectral-spatial details, intermediate layers integrating local context, and deep layers fusing global relationships-all guided by prompt information injected at appropriate scales. This multi-resolution approach enables simultaneous preservation of small objects and coherence of large regions, a trade-off that pure transformers or CNNs typically struggle to achieve.

Across datasets, VP-Hype yields sharper boundaries, stronger spatial coherence, and better preservation of thin and small regions. This follows from tight architectural coupling: a convolutional front end preserves local spectral-spatial priors; the Hybrid-Mamba backbone alternates linear-time state-space blocks for global context with windowed self-attention for local refinement; frozen CLIP textual prompts provide semantic priors while learnable visual prompts encode spatial templates; and TCSP cross-attention with learnable temperature plus prompt-fusion blocks inject a calibrated, multi-scale signal into the backbone. In short, textual prompts direct what to attend to, visual prompts indicate where to look, and the hybrid backbone determines how to integrate those cues, yielding sample-efficient, structurally reliable maps under extreme label sparsity.




\subsection{Prompt System Effectiveness}

Ablation studies on the Salinas dataset with 2\% training samples systematically 
isolate the contribution of each prompt modality within the VP-Hype framework. 
As reported in Table~\ref{tab:prompt-number-effect}, removing both prompt types 
yields the lowest performance (OA = 98.28\%, AA = 99.05\%, Kappa = 98.08\%), 
confirming that the prompt system is a critical component of the overall architecture.

Introducing textual prompts alone (4 prompts) produces a substantial gain of 
$+1.03\%$ in OA (99.31\%), $+0.49\%$ in AA (99.54\%), and $+1.15\%$ in Kappa 
(99.23\%), demonstrating that semantic guidance encoded through CLIP-based text 
embeddings has a strong regularizing effect on feature discrimination. Visual 
prompts alone (4 prompts) also improve over the baseline ($+0.54\%$ OA, 
$+0.26\%$ AA, $+0.60\%$ Kappa), albeit to a lesser degree, indicating that 
spatial pattern encoding provides complementary but secondary discriminative cues. 
Notably, textual prompts exert a greater influence on overall accuracy and class 
separability than visual prompts, suggesting that semantic context is the dominant 
inductive bias in this setting.

\begin{table}[h]
  \centering
  \small
  \setlength{\tabcolsep}{0pt}
  \renewcommand{\arraystretch}{1.2}
  \caption{Ablation study on the influence of the number of textual and visual prompts on VP-Hype performance (Salinas Dataset, 2\% training samples).}
  \label{tab:prompt-number-effect}
  \begin{tabular*}{\columnwidth}{@{\extracolsep{\fill}} cc ccc @{}}
    \toprule
    \textbf{Textual Prompts} & \textbf{Visual Prompts} & \textbf{OA (\%)} & \textbf{AA (\%)} & \textbf{Kappa (\%)} \\
    \midrule
    None (Baseline) & None & 98.28 & 99.05 & 98.08 \\
    4 Prompts       & None & 99.31 & 99.54 & 99.23 \\
    None            & 4 Prompts & 98.82 & 99.31 & 98.68 \\
    2 Prompts       & 2 Prompts & 99.43 & 99.60 & 99.36 \\
    \textbf{4 Prompts}& \textbf{4 Prompts} & \textbf{99.69} & \textbf{99.78} & \textbf{99.65} \\
    \bottomrule
  \end{tabular*}
\end{table}

Combining both modalities (2 textual $+$ 2 visual prompts) achieves a higher OA 
of 99.43\% ($+1.15\%$ over baseline), already surpassing the individual uni-modal 
configurations, which evidence a clear synergistic interaction between spatial 
and semantic cues. Scaling to the full configuration (4 textual $+$ 4 visual 
prompts) yields the best performance across all metrics 
(OA = 99.69\%, AA = 99.78\%, Kappa = 99.65\%), representing gains of $+1.41\%$, 
$+0.73\%$, and $+1.57\%$ in OA, AA, and Kappa, respectively. This consistent 
monotonic improvement with prompt count confirms that richer prompt representations 
lead to more informative task-aware conditioning.

These results validate the design of the TCSP module, whose cross-attention 
mechanism with learnable temperature $\tau$ adaptively regulates the fusion 
strength between visual and textual cues, enabling the model to exploit 
complementary information without over-weighting either modality. The use of 
pre-computed CLIP embeddings ensure that this task-aware adaptation incurs no 
additional inference overhead, preserving computational efficiency while 
delivering consistent accuracy improvements across varying training regimes.

\section{Conclusion}
\label{sec:conclusion}
We present VP-Hype, a hybrid Mamba-Transformer architecture enhanced with visual-textual prompting for hyperspectral image classification. The approach synergistically combines state-space models' linear complexity with self-attention's representational capacity, augmented by a task-aware multi-modal prompt system that fuses learnable spatial patterns with CLIP-based semantic embeddings.

The hybrid architecture addresses fundamental challenges: Mamba's linear complexity enables efficient processing of high-dimensional spectral data while maintaining long-range dependencies, while attention captures fine-grained spatial-spectral relationships. The visual-textual prompt system enables task-aware adaptation by combining learnable visual templates with semantic text guidance, proving particularly effective in few-shot scenarios. Experiments demonstrate complementary benefits from visual and textual components, with cross-attention fusion yielding superior performance. The architecture demonstrates strong generalizability across diverse hyperspectral datasets, consistently achieving state-of-the-art performance under both standard and extreme data-scarcity conditions.

Limitations include fixed text descriptions that may not capture full diversity across acquisition conditions, computational efficiency that could be optimized for real-time applications, and a focus on classification tasks that could be extended to anomaly detection, target detection, and spectral unmixing. The method assumes known task categories, limiting effectiveness in open-set scenarios.

Future directions include developing adaptive prompt generation mechanisms, exploring efficient architecture variants for resource-constrained environments, investigating additional modalities such as temporal information, extending to open-set classification, and developing a theoretical understanding of multi-modal prompt fusion mechanisms.

\appendix

\section{Overall VP-Hype Framework}
The VP-Hype framework, detailed in Algorithm \ref{alg:vphype}, introduces a hybrid Mamba–Transformer architecture designed to capture the complex spatial-spectral dependencies inherent in hyperspectral imagery. The pipeline begins by transforming the input cube into patch embeddings, which are then enriched by CLIP-based textual prompts to provide task-specific semantic context. Within each hierarchical stage, the model balances global modeling efficiency and local feature refinement by alternating between MambaVisionMixer blocks and Windowed Self-Attention. A core innovation of this approach is the Textual-Conditioned Soft Prompting (TCSP) module, which dynamically fuses textual priors into the visual feature maps to enhance class discriminability. The process concludes with global aggregation and a linear classification layer to generate the final predictions.
\label{app:algorithm}

\begin{algorithm}[htp]
\caption{VP-Hype: Hybrid Mamba-Transformer with Visual-Textual Prompting}
\label{alg:vphype}
\begin{algorithmic}[1]
\REQUIRE Hyperspectral image $\mathbf{I}\in\mathbb{R}^{B\times C\times H\times W}$,
task identifier $t$ (optional),
CLIP text embeddings $\mathbf{E}_{clip}$,
learnable parameters $\Theta$
\ENSURE Class logits $\hat{\mathbf{Y}}\in\mathbb{R}^{B\times N}$

\STATE \textbf{Patch embedding:}
\STATE $\mathbf{X}_0 \leftarrow \mathrm{PatchEmbed}(\mathbf{I})$ \hfill $\triangleright$ Eq.~\eqref{eq:embed}

\STATE \textbf{Textual prompt encoding:}
\STATE $\mathbf{w} \leftarrow \mathrm{OneHot}(t)$
\STATE $\mathbf{e}_{t} \leftarrow \sum_{i=0}^{T-1} \mathbf{w}_{:,i}\odot\mathbf{E}_{clip}[i,:]$ \hfill $\triangleright$ Eq.~\eqref{eq:text_prompt}

\FOR{$l = 0$ \TO $L-1$}
    \STATE $\mathbf{X}_l \leftarrow \mathrm{StageInput}(\mathbf{X}_{l})$

    \FOR{$i = 0$ \TO $\text{depth}_l-1$}
        \IF{$i < \lfloor \text{depth}_l/2 \rfloor$}
            \STATE $\mathbf{X}_l \leftarrow \mathrm{MambaVisionMixer}(\mathbf{X}_l)$
        \ELSE
            \STATE $\mathbf{X}_l \leftarrow \mathrm{WindowedSelfAttention}(\mathbf{X}_l)$
        \ENDIF
        \STATE $\mathbf{X}_l \leftarrow \mathrm{MLP}(\mathbf{X}_l)$
    \ENDFOR

    \IF{stage $l$ uses prompting}
        \STATE \textbf{TCSP prompt generation:}
        \STATE $\mathbf{P}_l \leftarrow \mathrm{TCSP}(\mathbf{e}_{t}, \mathbf{P}_{v})$ \hfill $\triangleright$ Eq.~\eqref{eq:app_TCSP}
        \STATE \textbf{Prompt fusion:}
        \STATE $\mathbf{X}_l \leftarrow \mathrm{Fuse}(\mathbf{X}_l, \mathbf{P}_l)$ \hfill $\triangleright$ Eq.~\eqref{eq:fusion}
    \ENDIF

    \IF{$l < L-1$}
        \STATE $\mathbf{X}_{l+1} \leftarrow \mathrm{Downsample}(\mathbf{X}_l)$
    \ENDIF
\ENDFOR

\STATE \textbf{Global aggregation:}
\STATE $\mathbf{f}_{pool} \leftarrow \mathrm{GlobalAvgPool}(\mathbf{X}_{L-1})$ \hfill $\triangleright$ Eq.~\eqref{eq:pool}

\STATE \textbf{Classification:}
\STATE $\hat{\mathbf{Y}} \leftarrow \mathbf{W}_{cls}\mathbf{f}_{pool} + \mathbf{b}_{cls}$ \hfill $\triangleright$ Eq.~\eqref{eq:logits}

\RETURN $\hat{\mathbf{Y}}$
\end{algorithmic}
\end{algorithm}

\section{Model Size Comparison}
\renewcommand{\thetable}{B.\arabic{table}}
\setcounter{table}{0}

Parameter count provides an informative view of model scale and complexity. 
Table~\ref{tab:model_size} situates VP-Hype among representative baselines, 
ranging from lightweight designs with only hundreds of thousands of parameters 
to very large transformer models with hundreds of millions of parameters. VP-Hype occupies a 
deliberate middle ground: substantially more compact than extremely large 
transformers such as ViT-Huge, yet considerably more expressive than 
lightweight architectures such as Madanet or LANet. This balance is particularly 
significant given that VP-Hype's additional capacity is dedicated to the 
hybrid Mamba-Transformer backbone and the dual-modal prompt fusion module, 
both of which directly contribute to its superior classification performance 
rather than representing redundant over-parameterisation.

\begin{table}[htbp]
\centering
\small
\caption{Parameter counts of representative models, comparing compact CNNs, 
large-scale transformers, and the proposed VP-Hype, which achieves a 
favourable balance between model expressivity and computational footprint.}
\label{tab:model_size}
\begin{tabular}{lcc}
\toprule
\textbf{Model} & \textbf{Dataset / Setting} & \textbf{Params (M)} \\
\midrule
Madanet \cite{cui2023madanet}        & Salinas         & 0.16   \\
LANet \cite{Ding2021}                & Salinas         & 5.3    \\
PiT-B \cite{heo2021rethinking}       & Longkou (2\%)   & 73.8   \\
ViT-Huge \cite{Dosovitskiy2021}      & Longkou (2\%)   & 632.0  \\
\textbf{VP-Hype (ours)}              & All datasets    & \textbf{77.30} \\
\bottomrule
\end{tabular}
\end{table}

\section{Text Conditional Spatial Prompt (TCSP)}
\label{app:TCSP}

This appendix provides the detailed mathematical formulation of the Text Conditional  Spatial Prompt (TCSP) module used in the main architecture (Sec.~\ref{sec:method}). TCSP fuses task-conditioned textual cues with learnable visual prompts to generate spatially aligned guidance maps that are injected into the hybrid backbone.

\subsection{Textual prompt projection}
Let $\mathbf{E}_{clip}\in\mathbb{R}^{T\times512}$ denote the precomputed CLIP text embeddings for $T$ task descriptions, and let $\mathbf{w}\in\{0,1\}^{B\times T}$ be the one-hot encoding of task identifiers. The task-conditioned textual embedding is computed as follows:
\begin{equation}
\label{eq:app_text_prompt}
\mathbf{e}_{t} = \sum_{i=0}^{T-1} \mathbf{w}_{:,i} \odot \mathbf{E}_{clip}[i,:] \in \mathbb{R}^{B\times512}.
\end{equation}
This embedding is projected to the prompt dimension $d_p$ and expanded spatially:
\begin{equation}
\label{eq:app_text_expand}
\tilde{\mathbf{e}}_{t} = \mathbf{W}_{clip}\mathbf{e}_{t} \in \mathbb{R}^{B\times d_p}, \qquad
\mathbf{T} = \mathrm{Interp}\!\left(\mathrm{Reshape}(\tilde{\mathbf{e}}_{t};B,d_p,1,1), (S_p,S_p)\right) \in \mathbb{R}^{B\times d_p\times S_p\times S_p},
\end{equation}
where $\mathbf{W}_{clip}\in\mathbb{R}^{d_p\times512}$ is a learnable projection.

\subsection{Visual prompt parameterization}
Learnable visual prompts are represented by a compact spatial tensor
\begin{equation}
\label{eq:app_visual_prompt}
\mathbf{P}_{v} \in \mathbb{R}^{1\times d_p\times S_p\times S_p},
\end{equation}
which is broadcast across the batch dimension during training and inference.

\subsection{Cross-attention fusion}
The textual prompt $\mathbf{T}$ and the visual prompt $\mathbf{P}_{v}$ are fused via cross-attention. Queries are derived from the textual prompt, while keys and values are derived from the visual prompt:
\begin{align}
\label{eq:app_qkv}
\mathbf{Q} &= \mathcal{P}_{q}(\mathbf{T}), \\
\mathbf{K} &= \mathcal{P}_{k}(\mathbf{P}_{v}), \\
\mathbf{V} &= \mathcal{P}_{v}(\mathbf{P}_{v}),
\end{align}
where $\mathcal{P}_{q}$, $\mathcal{P}_{k}$, and $\mathcal{P}_{v}$ are learnable linear projections implemented as lightweight convolutions.

After flattening the spatial dimensions, cross-attention is computed as
\begin{equation}
\label{eq:app_attention}
\mathbf{A} = \mathrm{Softmax}\!\left(\frac{\mathbf{Q}\mathbf{K}^{\top}}{\sqrt{d_p}\,\tau}\right), \qquad
\mathbf{U} = \mathbf{A}\mathbf{V},
\end{equation}
where $\tau>0$ is a learnable temperature parameter controlling attention sharpness.

\subsection{Prompt refinement and output}
The attention output is reshaped back to spatial form and refined by a gated feed-forward network:
\begin{equation}
\label{eq:app_refine}
\mathbf{P}_{fused} = \mathrm{FFN}_{g}\!\left(\mathrm{Reshape}_{sp}(\mathbf{U})\right) \in \mathbb{R}^{B\times d_p\times S_p\times S_p}.
\end{equation}
Finally, the fused prompt is resized and projected to match the backbone feature geometry:
\begin{equation}
\label{eq:app_output}
\mathbf{P} = \mathbf{W}_{proj}\,\mathrm{Interp}\!\left(\mathbf{P}_{fused}, (H_f,W_f)\right) \in \mathbb{R}^{B\times C_f\times H_f\times W_f},
\end{equation}
where $\mathbf{W}_{proj}$ is implemented as a lightweight $3\times3$ convolution. For compact notation in the main text, the above operations are summarized as
\begin{equation}
\label{eq:app_TCSP}
\mathbf{P} = \mathrm{TCSP}(\mathbf{e}_{t}, \mathbf{P}_{v}).
\end{equation}

\begin{table}[h]
  \centering
  \scriptsize
  \setlength{\tabcolsep}{0pt}
  \renewcommand{\arraystretch}{1.1}
  \caption{Overall, average, kappa, and per-class accuracies (\%) on the Salinas dataset (10\% training samples).}
  \label{tab:salinas-10p-full}
  \begin{tabular*}{\columnwidth}{@{\extracolsep{\fill}} l cccccccc @{}}
    \toprule
    \textbf{Class} & \textbf{Ours} & LoLA & MADANet & A$^2$MFE & HybridSN & SSFTT & HiT & LANet \\
    \midrule
    OA (\%) & \textbf{99.99} & 99.87 & 99.17 & 98.56 & 97.05 & 98.61 & 97.83 & 96.67 \\
    AA (\%) & \textbf{99.99} & 99.71 & 99.12 & 98.72 & 97.52 & 98.97 & 98.87 & 97.12 \\
    Kappa (\%) & \textbf{99.99} & 99.49 & 98.34 & 97.35 & 96.72 & 98.46 & 97.58 & 96.39 \\
    \midrule
    Brocoli green weeds 1 & \textbf{100.00} & 100.00 & 99.64 & 99.46 & 99.30 & 99.34 & 100.00 & 98.32 \\
    Brocoli green weeds 2 & \textbf{100.00} & 100.00 & 98.95 & 97.28 & 99.96 & 100.00 & 99.75 & 98.64 \\
    Fallow & \textbf{100.00} & 100.00 & 98.90 & 97.99 & 98.16 & 99.89 & 100.00 & 95.09 \\
    Fallow rough plow & \textbf{100.00} & 100.00 & 97.58 & 95.67 & 98.05 & 99.34 & 99.85 & 98.01 \\
    Fallow smooth & \textbf{99.96}  & 99.59  & 99.02 & 98.85 & 98.42 & 99.50 & 97.59 & 98.73 \\
    Stubble & \textbf{100.00} & 100.00 & 98.01 & 98.31 & 98.67 & 98.31 & 99.92 & 98.86 \\
    Celery & \textbf{100.00} & 100.00 & 99.34 & 95.76 & 98.96 & 99.91 & 100.00 & 97.04 \\
    Grapes untrained & \textbf{100.00} & 100.00 & 97.77 & 97.82 & 92.70 & 97.80 & 95.81 & 96.49 \\
    Soil vinyard develop & \textbf{100.00} & 100.00 & 98.23 & 97.56 & 98.53 & 100.00 & 100.00 & 93.11 \\
    Corn senesced green weeds & \textbf{100.00} & 99.80 & 95.91 & 95.23 & 95.44 & 99.81 & 98.19 & 94.27 \\
    Lettuce romaine 4wk & \textbf{100.00} & 100.00 & 98.86 & 97.45 & 98.08 & 98.77 & 99.42 & 97.19 \\
    Lettuce romaine 5wk & \textbf{100.00} & 100.00 & 99.23 & 98.34 & 99.33 & 99.89 & 99.84 & 98.42 \\
    Lettuce romaine 6wk & \textbf{100.00} & 100.00 & 100.00 & 99.21 & 99.02 & 97.02 & 100.00 & 98.58 \\
    Lettuce romaine 7wk & \textbf{100.00} & 100.00 & 99.34 & 98.12 & 98.21 & 98.96 & 99.42 & 98.07 \\
    Vinyard untrained & \textbf{100.00} & 95.69  & 98.49 & 94.55 & 92.29 & 95.52 & 92.53 & 92.46 \\
    Vinyard vertical trellis & \textbf{100.00} & 100.00 & 98.36 & 93.23 & 80.43 & 99.49 & 99.66 & 90.11 \\
    \bottomrule
  \end{tabular*}
\end{table}

\begin{table}[]
  \centering
  \scriptsize
  \setlength{\tabcolsep}{0pt}
  \renewcommand{\arraystretch}{1.1}
  \caption{Overall, average, kappa, and per-class accuracies (\%) on the HongHu dataset (10\% training samples).}
  \label{tab:honghu-full}
  \begin{tabular*}{\columnwidth}{@{\extracolsep{\fill}} l ccccccc @{}}
    \toprule
    \textbf{Class} & \textbf{Ours} & HSIMAE & HybridSN & ViT & MASSFormer & DBCT & LSGA \\
    \midrule
    OA (\%)    & \textbf{99.64} & 96.24 & 96.76 & 96.76 & 98.14 & \underline{99.14} & 99.02 \\
    AA (\%)    & \textbf{99.23} & 92.99 & 92.50 & 92.50 & 96.25 & \underline{98.07} & 97.54 \\
    Kappa (\%) & \textbf{99.55} & 96.18 & 96.77 & 96.77 & 97.65 & \underline{98.91} & 98.75 \\
    \midrule
    Red roof                & \textbf{99.84} & 97.40 & 98.46 & 97.84 & 98.99 & 98.27 & \underline{99.54} \\
    Road                    & \textbf{99.71} & 88.87 & 75.45 & 87.22 & \underline{97.25} & 97.44 & 96.82 \\
    Bare soil               & \textbf{99.59} & 95.84 & 95.61 & 94.98 & 98.87 & \underline{99.01} & 98.23 \\
    Cotton                  & \textbf{100.00} & 98.13 & 99.40 & 98.48 & 99.31 & \underline{99.69} & 99.65 \\
    Cotton firewood         & \textbf{99.62} & 78.73 & 88.93 & 84.61 & 94.76 & 98.32 & \underline{99.14} \\
    Rape                    & \textbf{99.84} & 98.81 & 98.41 & 98.60 & 99.11 & \underline{99.47} & 99.34 \\
    Chinese cabbage         & \textbf{99.50} & 94.10 & 95.77 & 92.50 & 97.15 & \underline{98.81} & 98.01 \\
    Pakchoi                 & \textbf{99.23} & 87.53 & 93.72 & 87.50 & 94.16 & \underline{98.38} & 96.59 \\
    Cabbage                 & \textbf{99.82} & 98.14 & 97.91 & 97.91 & \underline{99.77} & 99.71 & 99.82 \\
    Tuber mustard           & \textbf{99.48} & 94.97 & 93.73 & 92.90 & 98.03 & 98.53 & \underline{99.94} \\
    Brassica parachinensis  & \textbf{99.55} & 91.95 & 91.88 & 88.10 & 95.53 & \underline{98.77} & 97.75 \\
    Brassica chinensis      & \textbf{98.46} & 92.80 & 92.17 & 91.23 & 98.15 & \underline{98.42} & 96.98 \\
    Small Brassica chinensis & \textbf{99.56} & 92.68 & 94.22 & 91.70 & 94.73 & \underline{98.48} & 97.63 \\
    Lactuca sativa          & \textbf{99.24} & 95.31 & 92.83 & 91.74 & 98.48 & \underline{99.02} & 98.90 \\
    Celtuce                 & \textbf{100.00} & 94.56 & 87.59 & 94.02 & 98.51 & \underline{98.18} & 97.96 \\
    Film covered lettuce    & \textbf{99.98} & 98.14 & 97.30 & 97.69 & 97.03 & 98.59 & \underline{99.16} \\
    Romaine lettuce         & \textbf{99.93} & 96.94 & 92.21 & 95.59 & 98.86 & \underline{99.47} & 99.32 \\
    Carrot                  & \textbf{96.86} & 90.37 & 89.72 & 93.51 & 95.90 & 95.16 & \underline{96.56} \\
    White radish            & \textbf{99.63} & 95.63 & 91.73 & 91.34 & 95.61 & 90.11 & \underline{99.51} \\
    Garlic sprout           & \textbf{99.68} & 88.67 & 89.22 & 86.62 & 93.85 & \underline{99.00} & 97.48 \\
    Broad bean              & \textbf{96.65} & 85.15 & 83.35 & 84.24 & 87.52 & \underline{90.77} & 91.62 \\
    Tree                    & \textbf{99.97} & 91.16 & 95.38 & 88.90 & 88.02 & \underline{96.76} & 98.08 \\
    \bottomrule
  \end{tabular*}
\end{table}

\bibliography{name}
\bibliographystyle{unsrt}

\end{document}